\documentclass[conference,a4paper,10pt]{IEEEtran}
\IEEEoverridecommandlockouts

\usepackage[hidelinks]{hyperref}
\usepackage[cmex10]{amsmath}%American Math Society(AMS) math formatting
\usepackage{amssymb,amsfonts}%AMS extra symbols and fonts
\usepackage{dblfloatfix}%fix double column figure ordering and placement

\usepackage[ruled,vlined]{algorithm2e}
\usepackage{graphicx}
\graphicspath{{Figures/PDF/}{Figures/PNG/}}
\usepackage{booktabs}
\usepackage{siunitx}
\usepackage[numbers,compress]{natbib}
\usepackage{subcaption}
\usepackage{texnames}
\usepackage{bm,bbm}
\usepackage{orcidlink}

\begin{document}

\title{\uppercase{Rapid Forest Fuel Load Estimation via Virtual Remote Sensing and Metric-Scale Feed-Forward 3D Reconstruction}}

\author{
    \IEEEauthorblockN{Quanyun Wu}
	\IEEEauthorblockA{
    \textit{University of Waterloo}\\
		Waterloo, Canada\\
		q34wu@uwaterloo.ca}
	\and
	\IEEEauthorblockN{Kyle Gao}
    	\IEEEauthorblockA{\textit{University of Waterloo}\\
    		Waterloo, Canada\\
    		y56gao@uwaterloo.ca}
	\and
    \IEEEauthorblockN{Wentao Sun}
    	\IEEEauthorblockA{\textit{University of Waterloo}\\
    		Waterloo, Canada\\
    		w27sun@uwaterloo.ca}

    \and
    \IEEEauthorblockN{Zhengsen Xun}
    	\IEEEauthorblockA{\textit{University of Calgary}\\
    		Calgary, Canada\\
    		zhengsen.xu@ucalgary.ca}

    \and
    \IEEEauthorblockN{Hudson Sun}
    	\IEEEauthorblockA{\textit{University of Waterloo}\\
    		Waterloo, Canada\\
    		h294sun@uwaterloo.ca}

    \and
    \IEEEauthorblockN{Linlin Xu}
    	\IEEEauthorblockA{\textit{University of Calgary}\\
    		Calgary, Canada\\
    		lincoln.xu@ucalgary.ca}
            
    \and
    \IEEEauthorblockN{Yuhao Chen}
    	\IEEEauthorblockA{\textit{University of Waterloo}\\
    		Waterloo, Canada\\
    		yuhaochen1@uwaterloo.ca}
    \and
    \IEEEauthorblockN{David A. Clausi}
    	\IEEEauthorblockA{\textit{University of Waterloo}\\
    		Waterloo, Canada\\
    		dclausi@uwaterloo.ca}
    \and
    \IEEEauthorblockN{Jonathan Li}
    	\IEEEauthorblockA{\textit{University of Waterloo}\\
    		Waterloo, Canada\\
    		junli@uwaterloo.ca}
    
}

\maketitle
\begin{abstract}
Accurate quantification of forest coverage and combustible biomass (fuel load) is critical for wildfire risk assessment and ecosystem management. However, traditional methods relying on airborne LiDAR or field surveys are cost-prohibitive and time-intensive, while satellite imagery often lacks the vertical resolution required for canopy volume analysis. This paper proposes a novel, automated pipeline for rapid forest inventory using virtual remote sensing data derived from Google Earth Studio (GES). Our approach first generates low-altitude orbital imagery and camera poses for a target region. For dense 3D reconstruction, we employ Pi-Long, developed within the VGGT-Long framework. This model serves as a scalable extension of the Pi-3 feed-forward Transformer architecture. To address the inherent scale ambiguity in monocular reconstruction, we introduce a metric recovery module that aligns the reconstructed trajectory with GES ground truth poses via Sim(3) Umeyama optimization. The metric-scale point cloud is then orthogonally projected into Bird’s-Eye-View (BEV) height and density maps. Finally, we employ a watershed-based segmentation algorithm combined with height variance analysis to classify tree species (conifer vs. broadleaf), calculate Leaf Area Index (LAI), and estimate total fuel load. Experimental results demonstrate that this pipeline offers a scalable, cost-effective alternative to physical scanning, enabling near-real-time estimation of forest biomass with high geometric consistency.

\end{abstract}


% ==========================================
% SECTION: INTRODUCTION
% ==========================================

The increasing frequency and intensity of wildfires globally have underscored the urgent need for precise, scalable tools to monitor forest fuel loads. Combustible biomass, defined as the total amount of live and dead vegetation available for fire consumption, is a primary variable in fire behavior models. Accurate estimation of parameters such as forest coverage area, canopy height, and species composition is essential for calculating fuel density and implementing mitigation strategies like prescribed burns.

Conventionally, forest inventory relies on manual field sampling or sampling using drone-based sensor platforms. While highly accurate, field surveys are labor-intensive and geographically limited. Airborne LiDAR and cameras provide excellent structural detail but involve high acquisition costs and complex logistics, making frequent temporal monitoring impractical for large-scale territories. Conversely, traditional passive satellite remote sensing offers broad coverage but often fails to capture the 3D structure of the canopy required for volumetric analysis.

To bridge the gap between high-cost aerial scanning and low-resolution satellite imagery, we propose a Virtual Remote Sensing paradigm utilizing Google Earth Studio (GES). While GES provides photorealistic imagery, converting these 2D image sequences into actionable metric data remains a challenge due to scale ambiguity and texture repetition in dense vegetation.

\begin{figure}[htbp]
\centering
\includegraphics[width=0.45\linewidth]{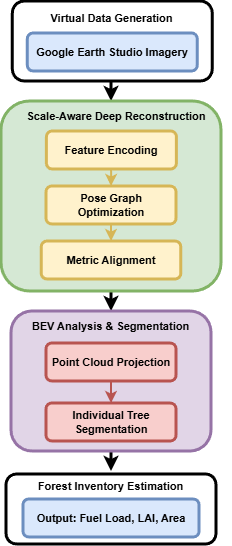} % Placeholder box
\caption{The proposed Virtual Remote Sensing pipeline. The workflow transitions from GES data acquisition to Pi-Long deep reconstruction, followed by Sim(3) metric scale recovery, and concludes with BEV projection for automated biomass estimation.}
\label{fig:pipeline}
\end{figure}

To address these challenges, we present an end-to-end framework as illustrated in Fig.~\ref{fig:pipeline}, that will allow forestry and wildfire researchers to access forest inventory estimations without new on-site data acquisition. Our methodology follows a strict four-stage sequence based on specific algorithmic modules:
\begin{enumerate}
    \item \textbf{Virtual Acquisition:} We first define a target forest in GES and synthesize a low-altitude orbital image sequence along with ground truth camera poses, simulating an idealized drone flight.
    \item \textbf{Deep 3D Reconstruction:} The image sequence is fed into Pi-Long~\cite{pilongcode2025}, a feed-forward Transformer network, which was developed within the VGGT-Long framework~\cite{deng2025vggt}. Unlike iterative optimization, Pi-Long robustly estimates relative camera poses and dense depth maps even in texture-poor forest regions.
    \item \textbf{Metric Recovery:} A crucial step in our pipeline is solving the monocular scale ambiguity. We utilize a Sim(3) optimization module (implemented in our metric recovery module) that rigidly aligns the reconstructed trajectory with the GES ground truth poses, restoring the point cloud to real-world metric dimensions.
    \item \textbf{Geometric Analysis:} The metric point cloud is orthogonally projected into 2D layers. Finally, a height-constrained watershed algorithm segments individual trees to compute forest area, LAI, and total fuel load.
\end{enumerate}
\section{Introduction}

% ==========================================
% SECTION: METHODOLOGY
% ==========================================
\section{Methodology}

The proposed framework enables rapid forest parametrization through a strictly sequential pipeline: Virtual Data Acquisition, Deep Sequential 3D Reconstruction, Metric Scale Recovery, and Automated Biomass Estimation.

\subsection{Virtual Data Acquisition via Google Earth Studio}
To bypass the logistical constraints of physical drone surveys, inspired by \cite{gao2025enhanced,Gao2025GBM}, we utilize Google Earth Studio (GES) as a virtual remote sensing image acquisition platform. This approach enables preliminary surveys without on-site data acquisition. It also supports simulated drone flights for testing drone-based image analysis frameworks. We define a spiral or circular flight path centered on the target forest coordinates. The system renders a sequence of high-resolution RGB frames ($I_{seq} = \{I_0, I_1, ..., I_N\}$) simulating a low-altitude aerial survey. Crucially, we simultaneously export the camera extrinsic parameters ($T_{GT}$) for each frame. While $I_{seq}$ serves as the sole input for the reconstruction network, $T_{GT}$ is reserved exclusively for the subsequent metric scale recovery phase.

\subsection{Feed-Forward 3D Reconstruction (Pi-Long)}
For dense 3D reconstruction, we employ Pi-Long, a scalable extension of the Pi-3~\cite{wang2025pi} feed-forward Transformer architecture. Unlike traditional SfM methods that rely on iterative bundle adjustment and are prone to feature matching failures in texture-repetitive vegetation, Pi-Long leverages deep geometric priors to infer camera poses and depth maps directly.

To handle long orbital sequences, we implement a sliding window strategy. The network predicts relative camera poses $T_{pred} \in SE(3)$ and depth maps for each local chunk. Subsequently, a loop closure optimizer detects overlapping visual features between non-adjacent chunks to minimize drift, generating a globally consistent trajectory $T_{recon}$ and a unified point cloud $P_{raw}$.

\subsection{Metric Scale Recovery via Sim(3) Optimization}
Monocular reconstruction is inherently scale-ambiguous (up-to-scale). To derive actionable forestry metrics, we must recover the absolute metric scale. We formulate this as a trajectory alignment problem between the reconstructed camera centers $C_{recon}$ and the ground truth camera centers $C_{GT}$ from GES. Using the Umeyama algorithm~\cite{umeyama1991least}, we compute the optimaltranslation, rotation, and scale $(t, R, s)$ that minimizes the least-squares error:
\begin{equation}
\min_{s, R, t} \sum_{i} || C_{GT}^i - (s R C_{recon}^i + t) ||^2.
\label{eq:umeyama}
\end{equation}
The computed combined transformation $(t, R, s) \rightarrow H_{sim3}$ is applied to $P_{raw}$ to generate the metric-scale point cloud $P_{metric}$, ensuring geometric dimensions match real-world measurements.

\subsection{BEV Projection and Biomass Estimation}
The 3D metric point cloud is converted into 2D analytical layers for fuel load estimation.

\subsubsection{Orthogonal Projection and Canopy Definition}
We apply PCA leveling and project the point cloud orthogonally to generate a normalized Height Map ($M_{height}$). Unlike traditional methods that rely on independent binary maps, we derive the Canopy Footprint Mask ($M_{canopy}$) directly from the height data using a robust relative elevation threshold ($h > h_{min}$, typically $0.15$). This ensures that ground noise is effectively filtered while preserving low-hanging branches.

\subsubsection{Physically Consistent Segmentation}
To address the "over-segmentation" problem common in dense forests, we propose a two-stage watershed approach, a standard methodology for canopy height model processing as described by Chen \textit{et al.} \cite{chen2006isolating}:
\begin{enumerate}
    \item \textbf{Tree Core Extraction:} We calculate global height statistics (mean $\mu_h$ and standard deviation $\sigma_h$) within the canopy footprint. A "Tree Core" mask is generated using an adaptive statistical threshold ($T_{core} = \mu_h + \alpha \cdot \sigma_h$), which isolates the high-confidence peaks of individual trees.
    \item \textbf{Constrained Watershed:} A Euclidean distance transform is applied specifically to these tree cores to find local maxima markers. The Watershed algorithm then expands these markers, strictly constrained within the $M_{canopy}$ boundaries.
\end{enumerate}

\subsubsection{Fuel Load}
% For each segmented tree $j$, we compute geometric attributes for species classification. Following the geometric statistical analysis proposed by Holmgren and Persson \cite{holmgren2004identifying}, we utilize the height variance within each segmented crown as a discriminator; trees exhibiting high verticality and low standard deviation ($\mu_h > 0.6 h_{max}, \sigma_h < \delta_{thresh}$) are classified as Conifers, while those with rounded crowns are classified as Broadleaf.

To ensure mass conservation, individual tree areas $A_j$ are scaled such that their sum equals the physical total canopy footprint ($\sum A_j = Area_{footprint}$). The total fuel load ($W_{fuel}$) is derived using allometric associations consistent with national biomass estimators \cite{jenkins2003national}:
\begin{equation}
W_{fuel} = \sum_{j} (A_{j}^{corrected} \times \alpha_{geo} \times LAI_{sp} \times \rho_{sp})
\label{eq:fuel}
\end{equation}
where $\alpha_{geo}$ is a geometric scaling factor based on latitude, $LAI_{sp}$ is a per-species leaf area index, and $\rho_{sp}$ is the per-species fuel sensitivity coefficient. Following Asner et al.~\cite{asner2003global}, we broadly group trees as broadleaf vs. coniferous, and set $\rho_{sp}$ to $3.8$ for broadleafs and $2.5$ for conifers, and we adopted $\mathrm{LAI}_{\mathrm{broadleaf}} = 5.5$ and $\mathrm{LAI}_{\mathrm{conifer}} = 3.0$. We introduced a latitude-dependent geometric correction factor, $\alpha_{geo}$, to account for macro-scale variations in canopy architecture. As noted by Chen et al.~\cite{chen2005derivation}, high-latitude coniferous forests are characterized by needle-shaped leaves that are tightly grouped into shoots rather than uniformly distributed. This structure creates significant gaps within the crown that are often indistinguishable in top-down imagery. Consequently, the raw projected area tends to overestimate the actual effective leaf area; thus, we apply a reduction factor of 0.85 for latitudes $latitude \ge 50^{\circ}$. Conversely, for tropical regions ($latitude < 23.5^{\circ}$), we applied a scaling factor of 1.15 to account for multi-layered canopy stratification following~\cite{asner2003global}, where substantial biomass exists beneath the upper canopy envelope visible to the camera.

\section{Result}
To validate the generalizability of our pipeline, we evaluated it on two distinct forest stands with contrasting canopy structures: (1) a Broadleaf-dominated mixed forest characterized by continuous, overlapping crowns, and (2) a Coniferous monoculture characterized by discrete, conical tree shapes.

\subsection{Qualitative 3D Reconstruction}
\begin{figure}[htbp]
\centering
% --- Row 1: Broadleaf ---
\begin{subfigure}[t]{0.45\linewidth}
  \centering
  \includegraphics[width=\linewidth]{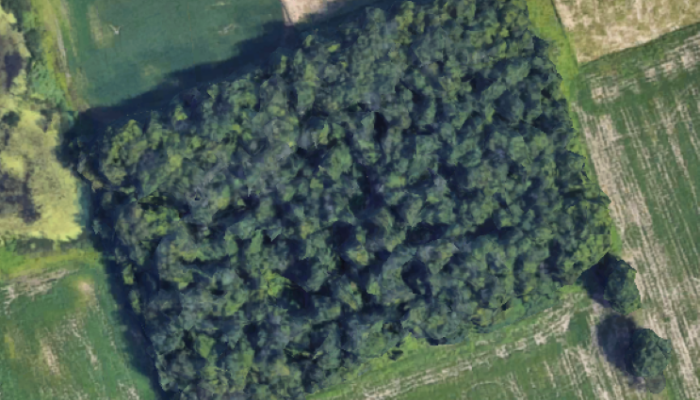}
  \caption{Broadleaf: GES Input}
  \label{fig:ges_broadleaf}
\end{subfigure}
\begin{subfigure}[t]{0.45\linewidth}
  \centering
  \includegraphics[width=\linewidth]{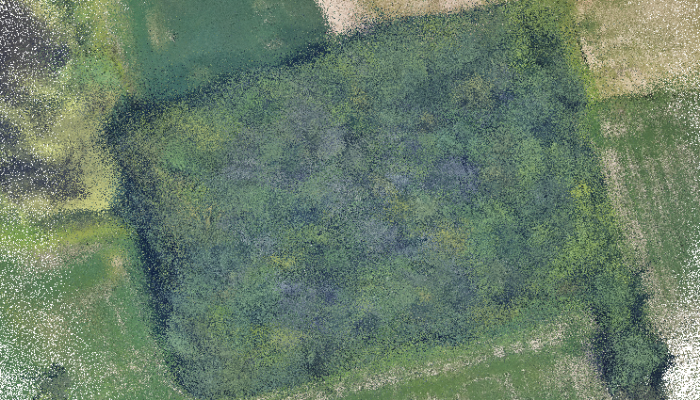}
  \caption{Broadleaf: 3D Reconstruction}
  \label{fig:recon_broadleaf}
\end{subfigure}

\vspace{0.1cm}

% --- Row 2: Conifer ---
\begin{subfigure}[t]{0.45\linewidth}
  \centering
  \includegraphics[width=\linewidth]{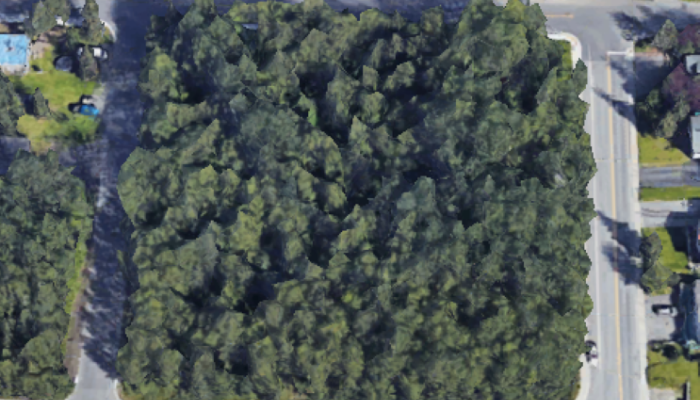}
  \caption{Conifer: 3D GES Input}
  \label{fig:ges_conifer}
\end{subfigure}
\begin{subfigure}[t]{0.45\linewidth}
  \centering
  \includegraphics[width=\linewidth]{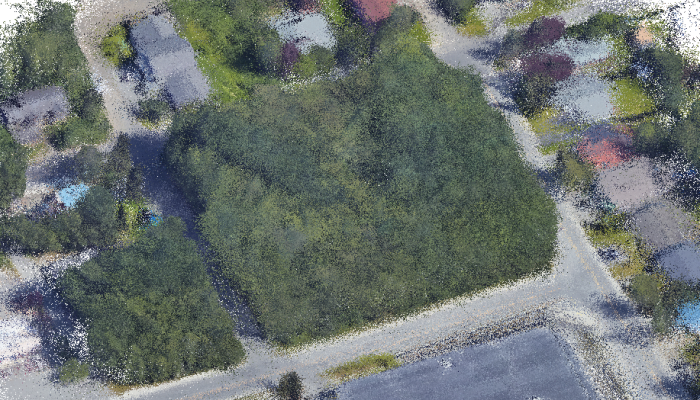}
  \caption{Conifer: 3D Reconstruction}
  \label{fig:recon_conifer}
\end{subfigure}
\caption{Qualitative evaluation across diverse forest types. The top row demonstrates the reconstruction of a closed-canopy Broadleaf stand, where the algorithm successfully recovers continuous crown surfaces. The bottom row depicts a Coniferous stand, highlighting the model's ability to resolve discrete, spike-like tree structures from the same orbital viewing angles.}
\label{fig:3d_comparison}
\end{figure}

Fig. \ref{fig:3d_comparison} compares the reconstruction fidelity across both datasets. In the broadleaf scenario (Fig. \ref{fig:3d_comparison}a-b), Pi-Long successfully recovered the smooth, rolling topography of the canopy surface, handling the texture repetition inherent in dense foliage. Crucially, in the coniferous scenario (Fig. \ref{fig:3d_comparison}c-d), the pipeline resolved the high-frequency geometric details of individual tree spikes. This confirms that the metric recovery module remains robust regardless of the underlying biological structure, maintaining scale consistency for both spreading and conical geometries.

% Figure 2: GES Screenshot & Point Cloud
% \begin{figure}[htbp]
% \centering
% \includegraphics[width=0.45\linewidth, height=3cm]{Figures/PNG/GES.png} \hfill % Placeholder for GES
% \includegraphics[width=0.45\linewidth, height=3cm]{Figures/PNG/PI-LONG.png}        % Placeholder for Point Cloud
% \caption{Visual comparison of the target forest. (a) The reference view from Google Earth Studio (GES). (b) The metric-scale dense point cloud reconstructed by Pi-Long.}
% \label{fig:reconstruction}
% \end{figure}

\subsection{BEV Projection and Segmentation}
\begin{figure}[htbp]
\centering
% --- Row 1: Broadleaf ---
\begin{subfigure}[t]{0.45\linewidth}
  \centering
  \includegraphics[width=\linewidth]{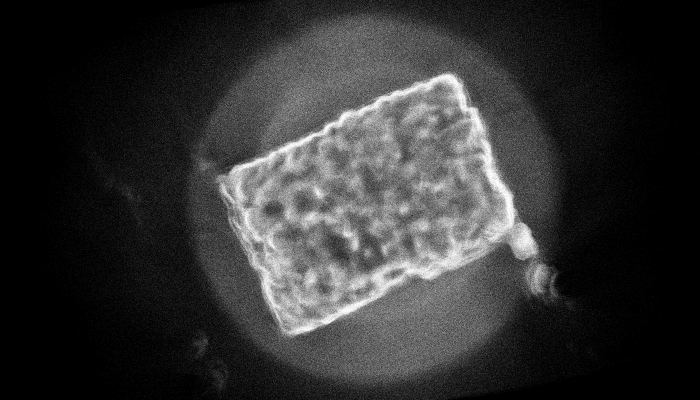}
  \caption{Broadleaf: Footprint}
  \label{fig:broadleaf_footprint}
\end{subfigure}
\begin{subfigure}[t]{0.45\linewidth}
  \centering
  \includegraphics[width=\linewidth]{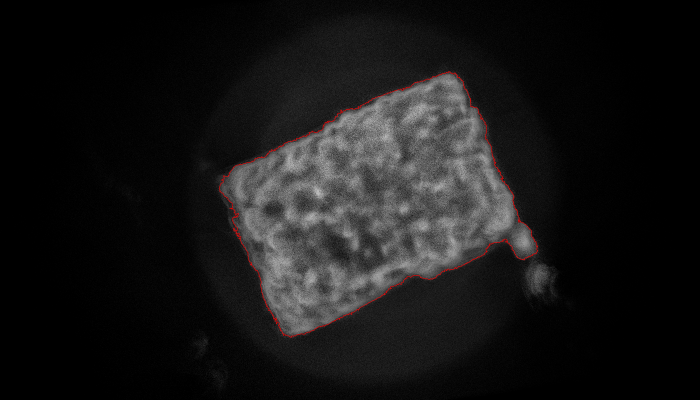}
  \caption{Broadleaf: Segmentation}
  \label{fig:broadleaf_seg}
\end{subfigure}
\vspace{0.1cm}
% --- Row 2: Conifer ---
\begin{subfigure}[t]{0.45\linewidth}
  \centering
  \includegraphics[width=\linewidth]{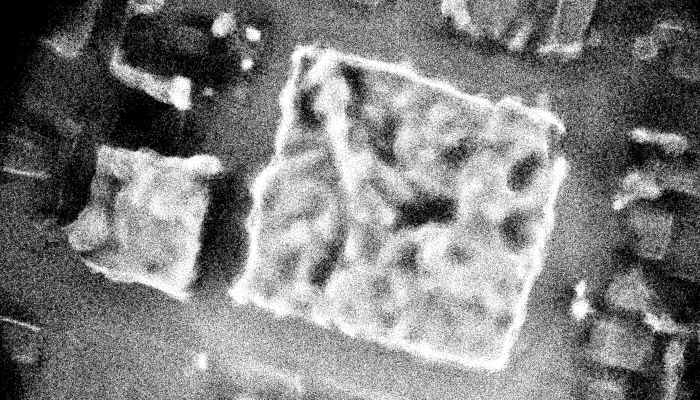}
  \caption{Conifer: Footprint}
  \label{fig:conifer_footprint}
\end{subfigure}
\begin{subfigure}[t]{0.45\linewidth}
  \centering
  \includegraphics[width=\linewidth]{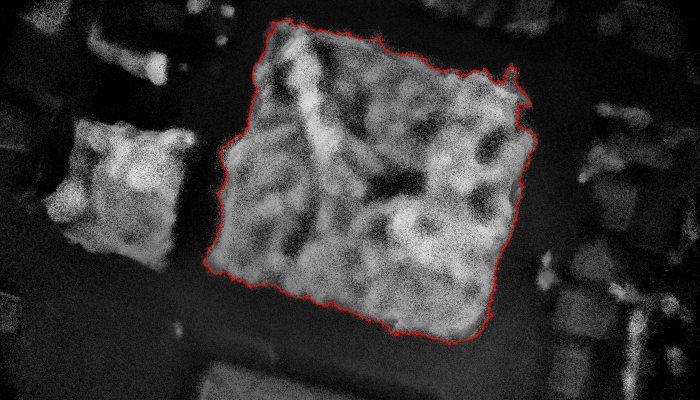}
  \caption{Conifer: Segmentation}
  \label{fig:conifer_seg}
\end{subfigure}
\caption{Comparative 2D analysis. (a-b) The broadleaf stand exhibits large, irregular canopy footprints; the watershed algorithm effectively delineates these merging crowns. (c-d) The coniferous stand shows distinct, separated markers in the binary map. The segmentation successfully isolates individual tree peaks (red contours) based on the sharp height gradients characteristic of needleleaf species.}
\label{fig:bev_comparison}
\end{figure}

The adaptability of our segmentation algorithm is visualized in Fig. \ref{fig:bev_comparison}.
\textbf{Broadleaf Case (Fig. 3a-b):} The canopy footprint appears as a merged, continuous mask. Here, the distance-transform watershed logic was critical for separating connected crowns based on topological centers rather than deep gaps, which are often obscured in broadleaf canopies~\cite{chen2006isolating}.
\textbf{Conifer Case (Fig. 3c-d):} The binary map shows distinct, isolated clusters. The height map reveals sharper gradients ($\sigma_h > 0.2$), which facilitated easier extraction of tree cores. The algorithm correctly identified smaller, denser clusters of trees without under-segmentation, demonstrating the effectiveness of the height-constrained watershed approach for diverse species.

% Figure 3: BEV Images (Binary & Height)
% \begin{figure}[htbp]
% \centering
% \includegraphics[width=0.45\linewidth, height=3cm]{Figures/PNG/bev_binary.png} \hfill % 
% \includegraphics[width=0.45\linewidth, height=3cm]{Figures/PNG/forest_result.png}       % Placeholder for BEV Height
% \caption{Intermediate 2D projections generated by the pipeline. (a) Binary Occupancy Map showing canopy extent. (b) Visualization of the boundary segmentation. The 2D Canopy Footprint Mask ($M_{canopy}$) is derived from the height map. The effective biological boundary (red contour) is back-projected onto the 3D point cloud.}
% \label{fig:bev_results}
% \end{figure}

\subsection{Quantitative Biomass Estimation}
The results, corrected for latitude-specific LAI variations, are summarized in Table~\ref{tab:biomass}. A key advantage of our updated algorithm is the Area Consistency Correction, which ensured that the sum of individual tree crowns exactly matched the computed Canopy Footprint area (40,965.3 $m^2$). This geometric constraint prevents the overestimation of biomass often caused by pixel overlap in traditional raster analysis.

% TABLE 1: Quantitative Results (Updated Labels)
\begin{table}[htbp]
\caption{Comparative Forest Inventory Estimation}
\centering
\begin{tabular}{|l|c|c|}
\hline
\textbf{Parameter} & \textbf{Broadleaf Stand} & \textbf{Conifer Stand} \\
\hline
\hline
Detected Trees & 896 & 336 \\ 
\hline
Forest Area ($m^2$) & 40,965.3 & 10305.6 \\ 
\hline
Assigned LAI & 5.5 & 2.55 \\
\hline
\textbf{Fuel Load (tons)} & 856.17 & 65.7 \\ 
\hline
\end{tabular}
\label{tab:biomass}
\end{table}
% \begin{table}[htbp]
% \caption{Automated Forestry Parameter Estimation}
% \centering 
% \begin{tabular}{|l|c|}
% \hline
% \textbf{Parameter} & \textbf{Estimated Value} \\
% \hline
% \hline
% Detected Trees (Count) & 896 \\ 
% \hline
% Forest Area (Reference) & 40,965.3 $m^2$ \\
% \hline
% Forest LAI (Leaf Area Index) & 5.5 \\ 
% \hline
% \textbf{Estimated Fuel Load} & \textbf{856.17 tons} \\ 
% \hline
% \end{tabular}
% \label{tab:biomass}
% \end{table}

\section{Discussion}

The primary objective of this study was to create a pipeline for fast and efficient preliminary biomass estimation. Our findings offer several critical insights.

\subsection{Overcoming the Texture-Repetition Barrier}
A significant challenge in forestry photogrammetry is the "texture repetition" problem, where traditional Structure-from-Motion (SfM) pipelines (e.g., COLMAP~\cite{schonberger2016structure}) fail to match features across similar-looking tree crowns. Our observations indicate that the feed-forward design of the Pi-Long architecture fundamentally addresses this limitation. By leveraging deep geometric priors learned from extensive datasets rather than relying solely on local feature matching, the system maintains trajectory continuity even in dense, homogeneous canopy regions. This represents a distinct advancement over the state-of-the-art in monocular reconstruction, which typically struggles with the high-frequency noise of vegetation.

\subsection{The Criticality of Metric Recovery}
While deep learning provides robust reconstruction, it inherently lacks physical scale (scale ambiguity). The successful derivation of biomass estimates in our results is entirely attributable to the integration of the Sim(3) optimization module. As validated by the alignment errors, relying on visual cues alone yields a "perfectly shaped" but dimensionally meaningless model. The explicit alignment of the reconstructed trajectory with the GES ground truth effectively bridges the gap between computer vision and quantitative ecology. This confirms that virtual remote sensing must be treated as a geometric registration problem, not just an image synthesis task.

\subsection{Limitations and Future Research}
Although cost-effective, our approach should be interpreted relative to airborne LiDAR. The reconstructed point cloud is closer to a digital surface model that captures the upper canopy envelope, rather than a full LiDAR return. Because photogrammetry cannot reliably penetrate foliage gaps~\cite{white2013best}, it cannot directly recover stem density or trunk diameter, so fuel load is inferred from canopy height to biomass correlations. This is adequate for preliminary assessment, but it can be less precise in dense canopies. Evaluation is also limited by the lack of a dataset that jointly provides RGB drone imagery, a 3D scan, and forest inventory ground truth for tree count, forested area, and fuel load.

As a continuation of this research, our team is planning a survey of the Broadleaf Stand scene using drone flights with both camera and LiDAR sensors, and ground transects to create a benchmark dataset for evaluating the proposed framework and its extensions. As an extension of our framework, We will combine geographic priors with multiview imagery and apply pretrained segmentation models such as SAM \cite{kirillov2023segment} alongside vision-language large language models. This integration is expected to support automated per-tree type classification in complex mixed stands and to yield more accurate fuel load estimates in heterogeneous forests.

\section{Conclusion}

This paper presented an end-to-end automated framework for rapid forest inventory, synthesizing virtual remote sensing, deep learning-based 3D reconstruction, and geometric analysis. By utilizing Google Earth Studio as a data source, we eliminated the logistical and financial barriers associated with physical aerial surveys.

The core contribution of this work is the creation of a pipeline that integrates Pi-Long for robust sequential reconstruction with a rigorous Sim(3) metric recovery process. This combination allows for the transformation of qualitative RGB pixels into quantitative, metric-scale ecological 3D models, which are then converted into measurements of forest area and combustible fuel load. Our approach democratizes access to forest monitoring tools, enabling high-frequency temporal analysis of wildfire risks.

\small
\bibliographystyle{IEEEtranN}
\bibliography{references}

@article{Gao2025GBM,
  title     = {{Gaussian Building Mesh (GBM): Extract a building's 3D mesh with Google Earth and Gaussian Splatting}},
  author    = {Gao, Kyle and Lu, Dening and He, Hongjie and Li, Liangzhi and Xu, Linlin and Chapman, Michael A. and Li, Jonathan},
  journal   = {Remote Sensing Applications: Society and Environment},
  volume    = {40},
  pages     = {101807},
  year      = {2025},
  publisher = {Elsevier},
}

@article{gao2025enhanced,
  title={{Enhanced 3D Urban Scene Reconstruction and Point Cloud Densification using Gaussian Splatting and Google Earth Imagery}},
  author={Gao, Kyle and Lu, Dening and He, Hongjie and Xu, Linlin and Li, Jonathan and Gong, Zheng},
  journal={IEEE Transactions on Geoscience and Remote Sensing},
  year={2025},
  volume    = {63},
  publisher={IEEE}
}

@misc{pilongcode2025,
  title = {Pi-Long: Extending $\pi^3$'s Capabilities on Kilometer-scale with the Framework of VGGT-Long},
  author = {{VGGT-Long Authors} and {$\pi^3$ Authors}},
  howpublished = {\url{https://github.com/DengKaiCQ/Pi-Long}},
  year = {2025},
  note = {GitHub repository}
}

@article{umeyama1991least,
  title={Least-squares estimation of transformation parameters between two point patterns},
  author={Umeyama, Shinji},
  journal={IEEE Transactions on Pattern Analysis and Machine Intelligence},
  volume={13},
  number={4},
  pages={376--380},
  year={1991},
  publisher={IEEE}
}

@article{deng2025vggt,
  title={VGGT-Long: Chunk it, Loop it, Align it--Pushing VGGT's Limits on Kilometer-scale Long RGB Sequences},
  author={Deng, Kai and Ti, Zexin and Xu, Jiawei and Yang, Jian and Xie, Jin},
  journal={arXiv preprint arXiv:2507.16443},
  year={2025}
}

@article{wang2025pi,
  title={$\pi^3$: Permutation-Equivariant Visual Geometry Learning},
  author={Wang, Yifan and Zhou, Jianjun and Zhu, Haoyi and Chang, Wenzheng and Zhou, Yang and Li, Zizun and Chen, Junyi and Pang, Jiangmiao and Shen, Chunhua and He, Tong},
  journal={arXiv preprint arXiv:2507.13347},
  year={2025}
}

@article{asner2003global,
  title={Global synthesis of leaf area index observations: implications for ecological and remote sensing studies},
  author={Asner, Gregory P and Scurlock, Jonathan MO and A. Hicke, Jeffrey},
  journal={Global ecology and biogeography},
  volume={12},
  number={3},
  pages={191--205},
  year={2003},
  publisher={Wiley Online Library}
}

@article{chen2006isolating,
  title={Isolating individual trees in a savanna woodland using small footprint lidar data},
  author={Chen, Qi and Baldocchi, Dennis and Gong, Peng and Kelly, Maggi},
  journal={Photogrammetric Engineering \& Remote Sensing},
  volume={72},
  number={8},
  pages={923--932},
  year={2006},
  publisher={American Society for Photogrammetry and Remote Sensing}
}

@article{jenkins2003national,
  title={National-scale biomass estimators for United States tree species},
  author={Jenkins, Jennifer C and Chojnacky, David C and Heath, Linda S and Birdsey, Richard A},
  journal={Forest science},
  volume={49},
  number={1},
  pages={12--35},
  year={2003},
  publisher={Oxford University Press}
}

@inproceedings{schonberger2016structure,
  title={Structure-from-motion revisited},
  author={Schonberger, Johannes L and Frahm, Jan-Michael},
  booktitle={Proceedings of the IEEE conference on computer vision and pattern recognition},
  pages={4104--4113},
  year={2016}
}

@article{white2013best,
  title={Best practices for generating forest inventory attributes from airborne laser scanning data using an area-based approach},
  author={White, Joanne C and Wulder, Michael A and Varhola, Andres and Vastaranta, Mikko and Coops, Nicholas C and Cook, Ross D and Pitt, Doug and Woods, Murray},
  journal={The Forestry Chronicle},
  volume={89},
  number={6},
  pages={722--723},
  year={2013}
}

@article{chen2005derivation,
  title={Derivation of global clumping index from POLDER data: Algorithm and validation},
  author={Chen, Jing M and Menges, CH and Leblanc, Sylvain G},
  journal={IEEE Transactions on Geoscience and Remote Sensing},
  volume={43},
  number={8},
  pages={1886--1896},
  year={2005},
  publisher={IEEE}
}

@inproceedings{kirillov2023segment,
  title={Segment anything},
  author={Kirillov, Alexander and Mintun, Eric and Ravi, Nikhila and Mao, Hanzi and Rolland, Chloe and Gustafson, Laura and Xiao, Tete and Whitehead, Spencer and Berg, Alexander C and Lo, Wan-Yen and others},
  booktitle={Proceedings of the IEEE/CVF international conference on computer vision},
  pages={4015--4026},
  year={2023}
}

\end{document}